\documentclass{article}

  \PassOptionsToPackage{numbers, compress}{natbib}
\usepackage{etoolbox}
\usepackage[belowskip=-9pt,aboveskip=0pt]{caption}
\usepackage[final]{nips_2017}

\usepackage[utf8]{inputenc} 
\usepackage[T1]{fontenc}    
\usepackage{hyperref}       
\usepackage{url}            
\usepackage{booktabs}       
\usepackage{amsfonts}       
\usepackage{nicefrac}       
\usepackage{microtype}      
\usepackage{epsfig}
\usepackage{graphicx}
\usepackage{subcaption}
\usepackage{multirow}
\usepackage{amsmath}
\usepackage{amssymb}
\usepackage{wrapfig}
\graphicspath{{figures/}{../figures/}}

\hypersetup{
    colorlinks=true,
    linkcolor=blue,
    filecolor=magenta,      
    urlcolor=magenta,
}
 
\title{Wide Inference Network for Image Denoising via Learning Pixel-distribution Prior
}

\author{
  Peng Liu
   \\
  University of Florida\\
  \texttt{pliu1@ufl.edu} \\
    \And
    Ruogu Fang\\
    University of Florida \\
    \texttt{ruogu.fang@bme.ufl.edu} \\
}

\begin{document}

\maketitle
\begin{abstract}

We explore an innovative strategy for image denoising by using convolutional neural networks (CNN) to learn similar pixel-distribution features from noisy images. Many types of image noise follow a certain pixel-distribution in common, such as additive white Gaussian noise (AWGN). By increasing CNN's width with larger reception fields and more channels in each layer, 
CNNs can reveal the ability to extract more accurate pixel-distribution features. The key to our approach is a discovery that wider CNNs with more convolutions tend to learn the similar pixel-distribution features, which reveals a new strategy to solve low-level vision problems effectively that the inference mapping primarily relies on the priors behind the noise property instead of deeper CNNs with more stacked nonlinear layers. We evaluate our work, Wide inference Networks (WIN), on AWGN and demonstrate that by learning pixel-distribution features from images, WIN-based network consistently achieves significantly better performance than current state-of-the-art deep CNN-based methods in both quantitative and visual evaluations.  \textit{Code and models are available at \url{https://github.com/cswin/WIN}}.

\textbf{The Correction:} This work has been leading relevant researchers with embedding knowledge domain, such as image Prior, into their tasks. However, our work does not have the same generalization capability with other image denoisers. There is a code issue that makes our work may be regarded as entirely out the way of the correct research direction, and meantime, it is misleading other researchers somehow. In particular, a good denoiser should be able to handle different noise values even which are always following the Gaussian distribution. In our implementation, we simulated the training noise with randn('seed,' 0) function, which resulted in one single and same noise matrix added on all training image patches. This is the reason why our work would have a bad performance without using randn('seed',0).  However, this work may still be good study material for the ones who would like to work on image restoration with deep learning.   

Besides, this work is one investigation with the motivation that is how to solve image restoration problem with simple but very effective way, and moreover,  this long-going problem can reach a breakthrough. 

Lastly,  you may try to think about how to map the particular matrix to generative ones.   Then, you may have a significant innovation published. 

\end{abstract}

\section{Introduction}
\label{introduction}

Over the last decade,  deep convolutional 
neural networks (CNNs) have revolutionized high-level vision tasks such as visual recognition, motion analysis, and object segmentation 
\cite{ciregan2012multi,farabet2013learning,krizhevsky2012imagenet}. Recently, CNNs have also been applied to low-level vision tasks such as super-resolution (SR) \cite{Kim_2016_VDSR}, image denoising \cite{zhang2016beyond,mao2016image}, and compression artifacts reduction \cite{dong2015compression}. In these tasks, a CNN is typically trained with supervised learning to represent a function – a mapping from a low-quality observation to a latent high perceptual visual image, with the goal of representing a restorer either removing the various types of noise or minimizing the effects of different artifacts from a degraded image. 

``Deeper is better'' typically has been generally accepted as a design criterion for building more powerful CNNs. 
The deep CNNs, such as VGG \cite{simonyan2014very}, GoogleNet \cite{szegedy2015going}, and ResNet \cite{he2016deep}, indeed have achieved a series of breakthroughs in high-level vision tasks. These deep nets have increasing number of layers of 19, 22, and 152 convolution layers\footnote{Only convolution layers are counted for the depth of CNN in general.}, with top 5 error rate on ILSVRC of 7.3\%, 6.7\%, and 3.57\% respectively. 
However, in contrast to the dominating privilege of deep nets in high-level vision tasks, most of recent works in low-level vision domain, such as DnCNN \cite{zhang2016beyond} and RED-Net \cite{mao2016image}, with up to 20 and 30 layers respectively, have not yet shown remarkable advantages compared to early methods (see Table~\ref{tab:comparison}). Obviously, in low-level vision tasks, which typically emphasizing more pixel-level features, depth is not the key. 

The success of deep CNNs in high-level vision domain is essentially due to a complicated nonlinear approximate function, which is trained on a large amount of labeled data through stacked convolution and nonlinear layers (e.g., ReLU \cite{nair2010rectified}). 
In addition, deep CNNs naturally integrate low / mid / high level features \cite{zeiler2014visualizing} and the ``levels'' of features can be enriched by the number of stacked layers (depth). Nevertheless, the extracted high-level features are not the key in low-level vision tasks. Instead, priors can be an important factor, which can capture statistical regularities of pixel-level features.  
A prior over the image space, such as non-local similarity, can help to come up with a very good estimate of the actual ``undo'' function to compensate for or undo defects which degrade an image \cite{li2009markov}. In image denoising, a notable prior in term of various types of noise is the pixel distributions implied in the noisy images, most of which follow regular distributions (e.g., Gaussian). 


\begin{table}[ht]
\vspace{-1em}
  \caption{Comparison of the structures of the most of the recent deep CNNs \cite{zhang2016beyond,mao2016image} for image denoising  
and the average peak signal-to-noise ratio (PSNR) improvement over one of the state-of-the-art 
non CNN-based method: BM3D \cite{dabov2009bm3d}. The results are evaluated on the gray-scale BSD100 and BSD200 dataset \cite{martin2001database}. As one can see, the average elevated margins measured by PSNR have not even surpassed 1 dB yet. 
Moreover, by comparing the gains obtained on BSD100 and BSD200, one can see that the generalization capability of both deep networks \cite{zhang2016beyond,mao2016image} 
\textit{decreases significantly} as the number of unseen test images increases.}
  \label{tab:comparison}
  \centering
  \begin{tabular}{llllll}
    \toprule
    \cmidrule{1-6}
   Methods  &  \# Layers  &  \# Filters  &  Filter Size  &  Gain-BSD100 (dB)  & 
 Gain-BSD200 (dB) \\
    \midrule
    DnCNN \cite{zhang2016beyond} & 20 & 64 & $3\times3$ & 0.7  & 0.56     \\
    RED-Net \cite{mao2016image}  & 30 & 64 & $7\times7$ & 0.73 & 0.43      \\
     
    \bottomrule
  \end{tabular}
\end{table}

Based on the above analysis, in this paper, we propose a CNN-based framework that can effectively learn feature distribution from noisy observations to form a prior for image denoising task. Our model, termed a \textit{Wide Inference Networks} (WIN), can capture the pixel-level distribution information, a capability which narrow and deep networks lack. 

Here we introduce the concept of ``width'' of the networks indicating both the number of filters in each layers and the size of the filters, as both parameters reflect the representation capability in one layer, in contrast to ``depth'' which demonstrates the non-linear representation power over the layers.

Specifically, we demonstrate the effectiveness of WIN within only 5 wider convolution layers, termed \textit{WIN5}, in the main denoising task on additive white Gaussian noise (AWGN). The performance gain is impressive. The key to our proposed network architecture is to employ larger perceptions fields through wider and shallower networks with more concentrated convolutions to capture the prior image distribution from the noisy images, and yields better overall generalization power to new, unseen noisy images.

\section{Background}
\label{background}
In this section, we provide background on distributions of image noise and spatial feature, the deep CNN-based image denoising methods\cite{dabov2009bm3d,zhang2016beyond,mao2016image}, 
regularization techniques, and learning strategies for generalization. In the sequel, we show that CNNs can be modeled and constructed with stronger capability to extract the spatial feature distribution, which can then be used as a prior for image denoising.

\begin{figure*}[t]
\centering
  \includegraphics[width=\textwidth]{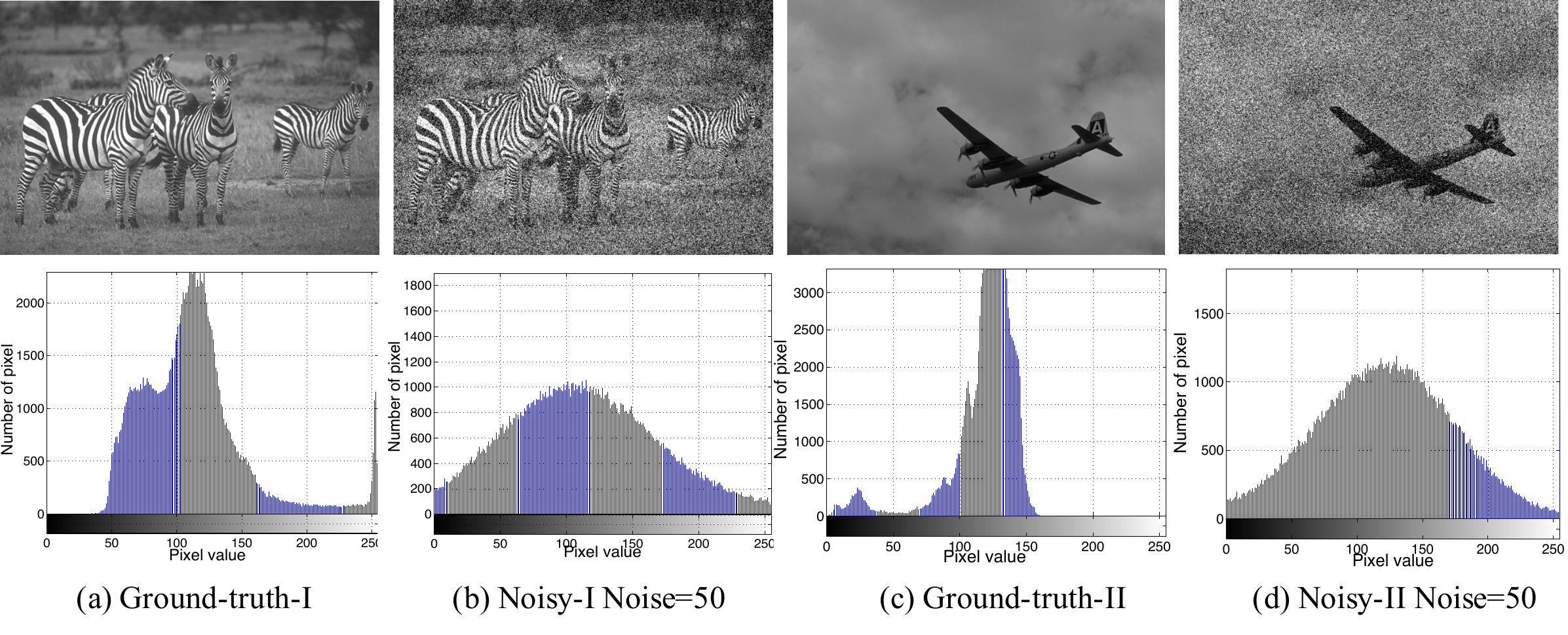}
  \caption{Similar distributions of histograms of two different
  images added additive white Gaussian noise (AWGN) with same noise level $\sigma=50$. }
  \vspace{-1em}
\label{fig:histogram_N50}
\end{figure*}

\textbf{Image Noise and Spatial Features Distribution:}
Image noise is pixel-level random variations and typically follows a certain distribution. 
Gaussian noise 
approximates a Gaussian distribution
and is usually used to mimic a realistic environment based on the Law of Large Numbers (LLN), which covers a very broad spectrum of practical applications. One common assumption is additive white Gaussian noise (AWGN) with different standard deviations, which can be formulated as $y_{i}=x_{i}+ n(size(x_{i}),\sigma)$, where $y$ is the noisy image, $x$ is the clean image, $\sigma$ is standard deviations of AWGN and determines the noise level. $n$ represents the Gaussian noise added to $x$ and essentially is a function that can return a matrix (with the same size of $x$) of Gaussian distributed random numbers, and $i$ is the index of the images from the dataset. From image histograms shown in Fig.~\ref{fig:histogram_N50}, one can see that no matter how different the features in $x_{i}$ are, as long as $\sigma$ and the sizes of $x_{i}$ are same, the different noisy images $y_{i}$ always have very similar pixel distributions. Such consistent representations are highly likely to be learned by CNNs. 

\textbf{Deep CNN-based Models:}
The deep
CNN-based state-of-the-art denoising models, DnCNN \cite{zhang2016beyond} and RED-Net \cite{mao2016image}, stem from the success of deep nets in high-level vision tasks \cite{simonyan2014very}. Particularly, DnCNN \cite{zhang2016beyond} adopts a 20 layers deep architecture, a learning strategy of residual learning \cite{kiku2013residual}, and a regularization method of batch normalization \cite{ioffe2015batch}. RED-Net \cite{mao2016image} employs a 30 layers deeper structure with skip connections (SK) added to connect corresponding layers of convolution to the deconvolution, and is justified by the residual network \cite{kiku2013residual}. In deep structures, learning strategies (e.g.residual learning \cite{kiku2013residual}) and regularization methods (e.g.batch normalization \cite{ioffe2015batch}) also work for accelerating the learning process and boosting performance. However, these models  
obtain good performance at the cost of growing complexity along with increasing network depth. Such strategy typically suffers from 
gradient vanishing, overfitting and degradation\footnote{Degradation refers to the reduction in accuracy with increasing depth of 
network after reaching a maxima.}.

 
\textbf{Learning Strategy: Residual learning
:}
Learning a residual representation is easier than estimating the desired objective directly since residual learning \cite{kiku2013residual} introduces more prior to the current objective.
Skip Connection is one form to introduce residual representation. 
Skip connection from input-end to output-end (input-to-output), like the one employed in VDSR \cite{Kim_2016_VDSR}, is able to compensate the lost details and perform residual learning simultaneously, formally, which holds $x_{i}=y_{i}+R(y_{i})$, where $(R(y_{i})\approx -n)$ is 
an embedded function for inferring residual--the opposite of noise added on $y_{i}$.
In RED-Net \cite{mao2016image}, the input is connected with the output to form a residual learning. In addition, there are connections every a few layers from convolutional feature maps to their mirrored deconvolutional maps to ease back-propagation, and reuse the otherwise lost details during deconvolution.
Another way to introduce residual learning is a mapping from an input observation to the corresponding \textit{precalculated} residual, which is adopted in DnCNN \cite{zhang2016beyond}. It aims to learn a mapping function $T(y_{i})\approx n$, and then it has $x_{i}=y_{i}-T(y_{i})$, 
where $x$, $y$, $i$ and $n$ are the same notation as aforementioned. Here $T$ is the objective output of DnCNN \cite{zhang2016beyond}, and $x_{i}$ is calculated separately after $T$ maps $y_{i}$ to the corresponding noise from the network.

\textbf{Regularization: Batch Normalization:}
One key to the success of DnCNN \cite{zhang2016beyond} is  
batch normalization (BN) \cite{ioffe2015batch}, which solves a problem called internal covariate shift, while accelerating network learning
and boosting accuracy. First of all, as data flow through a deep CNN, 
the weights and parameters adjust the output maps at each layers, some times leading to very large or small values in the intermediate feature maps. By normalizing the data in each mini-batch, this problem can be mostly avoided. Furthermore, BN also impacts gradient flow. Thus it can reduce the dependence on the scale of the parameters and the initial values, and prevent the network from getting stuck in saturated modes caused by certain non-linearities. 
\section{Wide Inference Network}
\label{ldnmodel}
The Wide Inference Network (WIN) is based on the plain convolutional neural network architecture with an exploration of the impact of the ``width'' of the network on the low-level vision tasks such as image denoising. In this section we introduce the investigation of the \textit{Wide Inference Network (WIN)} in three aspects: 
(1) How to represent and optimize WIN to find an efficient and high performing architecture 
that can learn the prior effectively; 
(2) The impact of residual learning \cite{kiku2013residual} and batch normalization
(BN) \cite{ioffe2015batch} when employed in \textit{WIN}. 
(3) The implementation details in the training stage. 
\begin{figure}[h!]
\centering
\vspace{-2em}
\includegraphics[width=\linewidth]{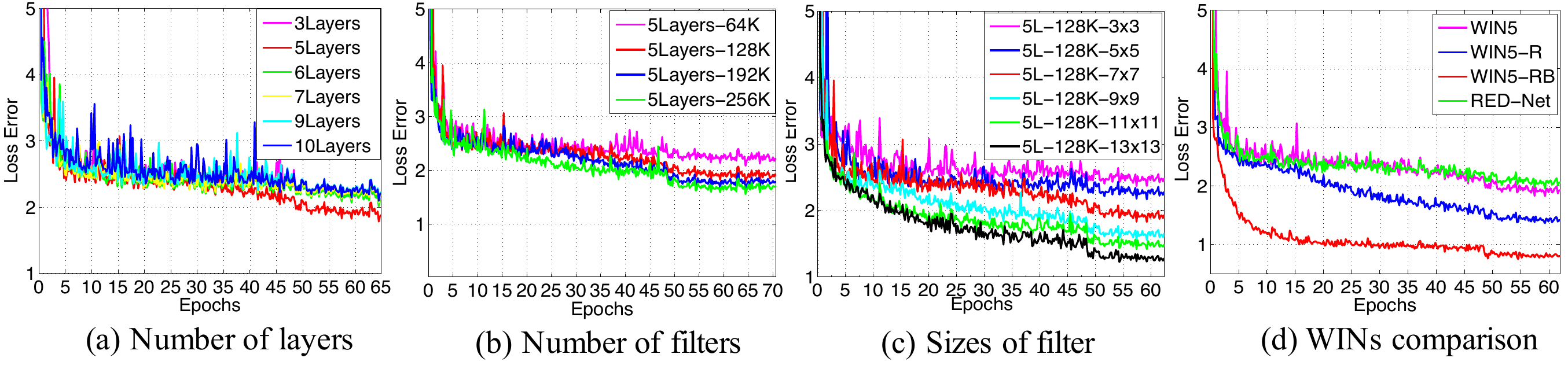}
   \caption{ Comparison of loss error on validation dataset during training for
   CNNs with different structure components: (a) Number of Layers; (b) Number of Filters; 
   (c) Size of Filters; (d) Proposed shallow wider nets with the state-of-the-art method RED-NET [16]. \textit{Note: Lower is better}.}
\label{fig:parameters}
\vspace{-2em}
\end{figure}
\subsection{Determining ``Width''}
In this section, we show an optimization process of a plain and wide denoising CNN to obtain a competitive model: \textit{WIN5}. In order to find the parameters that make the major contributions to performance improvement, we compare the loss errors on the same validation dataset during training from three groups of experiments of wide inference network with different number of layers, different number of filters for each layer, and different filter sizes respectively. The results are shown in Fig.~\ref{fig:parameters}. 
CNNs are made of a series of layers. The convolution (Conv) layer is the
core building block of a convolutional network that does most of the
computationally heavy lifting. Following the principle in \textit{Striving
for Simplicity: The All Convolutional Net} \cite{springenberg2014striving},
we build \textit{WIN5} with a sequence of Conv Layers without Pooling and fully connected (FC) layers, and each Conv layer is followed by a ReLU~\cite{nair2010rectified}, except for the last layer. A Conv layer is composed of a set of neurons with learnable weights and biases. The number of neurons and weights are referred to as the width of a CNN. Four hyper-parameters control the size of the width: the number of layers ($L$), the number of filters ($K$), the size of filters ($F$) and the input volume ($D$) of each Conv layer. The input volume of each layer is determined by the patch size of the input images (only for the first Conv
layer) or the output volume size of the previous Conv layer. We follow the common effective
settings of patch size \cite{zhang2016beyond} and keep the output volume of each layer the
same size as the patch size. Thus, we focus on comparing and analysing the main performance
factors: $L$, $K$ and $F$. The following sections compare the performance of several
WINs in the same experimental setting. All
comparing experiments are performed on a common data set BSD500 \cite{martin2001database} applied AWGN with noise level \(\sigma=50\).

\textbf{The Number of Layers ($L$).} As shown in Fig.~\ref{fig:parameters}(a), WIN with
the number of layers $L=5$ outperforms both shallower or deeper networks other than $L=5$. In addition, ``the deeper the worse'' (degradation) is apparent, which is caused
by the loss of image details through the deep network. Since all the filter weights in a convolutional network are learned, information loss is caused primarily by the output that CNN is mapped to. But the first few layers of the network usually learn small, local features and the network progressively  discriminatory elements as we go deeper \cite{zeiler2014visualizing}.

\textbf{The Number of Filters ($K$).} As we can see in Fig.~\ref{fig:parameters}(b), WIN with $K=128$ achieves remarkable performance gains than the network with $K=64$. $K=128$ is
found to be the optimal value for high performance in this denoising task. As $K$ in each layer increases, the performance
improves. However, as $K$ is more than $256$, the performance enhancement plateaus. Meantime, the corresponding training time and computation complexity always keep growing vastly. 

\textbf{The Size of Filters ($F$).} From Fig.~\ref{fig:parameters}(c), $F=7\times7$ is able
to improve performance remarkably compared to smaller $F$. In general, larger $F$ leads to
better performance. However, the performance improvement margin diminishes when $F$ is
greater than $7\times7$. Similar to the number of filters $K$, larger F also dramatically
increases both training time and computation complexity. 

To sum up, from the overall performance and efficiency point of view, WIN with
$L=5$, $K=128$, $F=7\times7$ is potentially the optimal denoising model among plain
shallow CNNs, and we refer it as WIN5 since it has 5 layers.  
Furthermore, WIN5 may achieve much more performance gains if we can cope 
with the degradation during training, as we address in the next Section~\ref{sec:degradation}

\subsection{Optimizing WIN5}
\subsubsection{WIN5-R: WIN5 + Residual Learning}
\label{sec:degradation}

Let us consider a Gaussian noisy observation $y=x+n$. Here, $y$ and $x$ are a noisy observation and the corresponding latent clean image. $n$ represents the Gaussian noise to be added to $x$.  DnCNN \cite{zhang2016beyond} aims to learn a mapping function \(T(y)\approx\displaystyle n\), and then it has \(x=y-T(y)\). For WIN5-R,
a skip connection from input-end to output-end is added to make up the lost details
and perform residual learning simultaneously, formally, which holds \(x=y+R(y)\)
where \(R(y)\approx\displaystyle -n\). In addition, as the end-to-end residual
learning needs to estimate the weights \(\Theta\) represented by the convolutions,
we minimize the Mean Squared Error (MSE) between noisy images (input \(y_{i}\))
and the clean versions as the ground-truth (label \(x_{i}\)) 
\begin{equation} \label{eq:1}
 l \left ( \Theta  \right )= \frac{1}{2N}\sum_{i=1}^{N}\left \| y_{i}+R\left ( y_{i};\Theta  \right ) -x_{i}\right \|_T^2 
\end{equation}
as the loss function to learn the trainable \(\Theta\). In Fig.~\ref{fig:parameters}(d),
we can see that WIN5-R (blue line) with a input-to-output skip connection can enhance the denoising performance by not only weakening degradation but also exploiting a faster and easier residual learning. 

\textbf{Comparison with RED-Net.} Fig.~\ref{fig:parameters}(d) provides the
comparison between our proposed shallow and wider nets with the state-of-the-art method RED-NET \cite{mao2016image} in terms of loss error. All our WIN-based architecture remarkably outperform RED-Net \cite{mao2016image} which is deeper and thinner.

\subsubsection{WIN5-RB: WIN5+Residual+Batch Normalization}

\begin{wrapfigure}{r}{0.5\textwidth}
 \vspace{-1em}
  \centering 
    \includegraphics[width=0.48\textwidth]{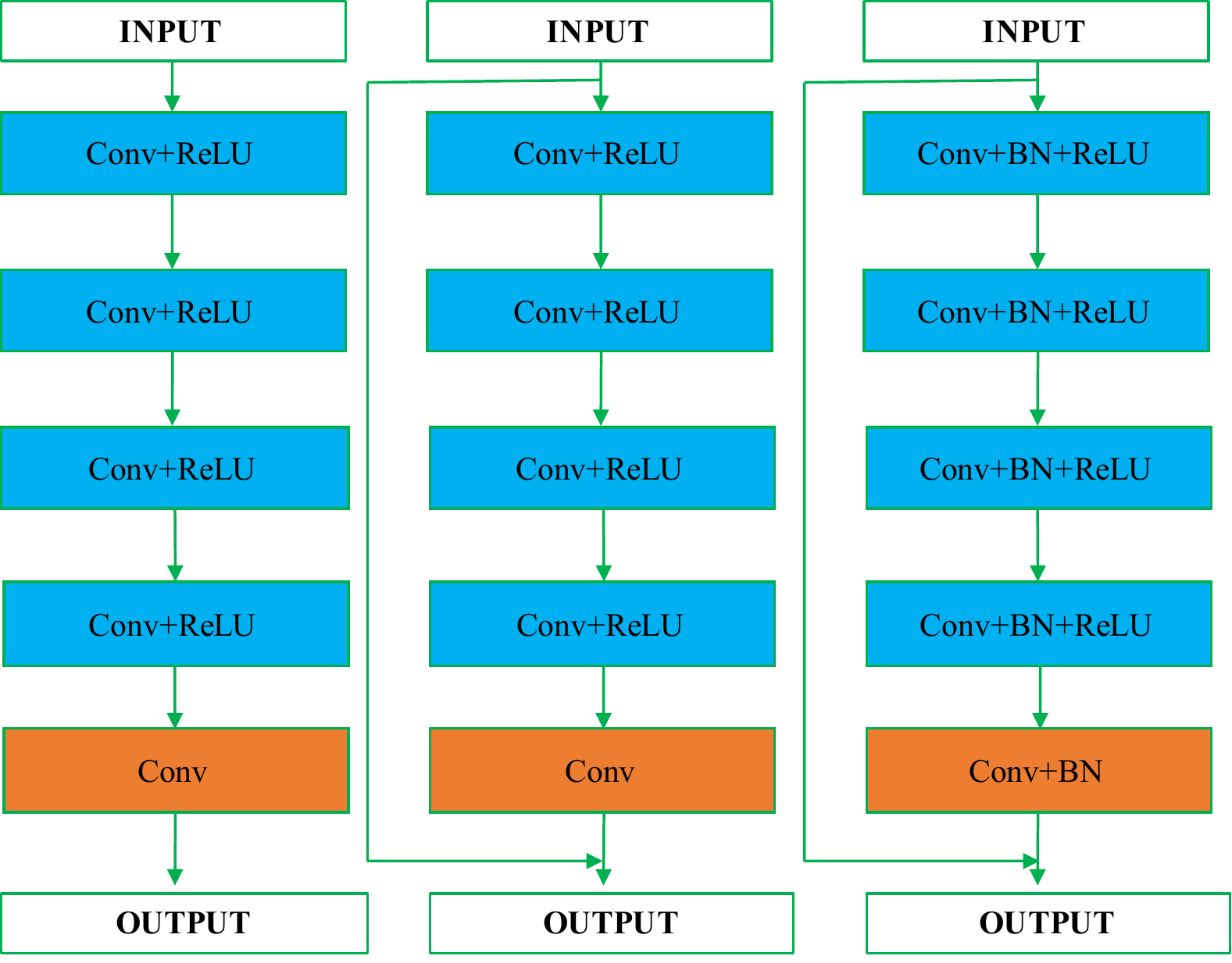}
  \caption{ Architectures (a) WIN5  (b) WIN5-R (c) WIN5-RB.}
  \label{fig:architecture}
  \vspace{-0em}
\end{wrapfigure}

In Fig.~\ref{fig:architecture}, we illustrate the evolutionary architectures of WIN5, WIN5-R, and WIN5-RB. 

\textbf{Architectures.} Three proposed models have the identical basic structure: $L=5$ layers and $K=128$ filters of size $F=7\times7$ in most convolution layers, except for the last one with $K=1$ filter of size $F=7\times7$. The differences among them are whether batch normalization (BN) and an input-to-output skip connection are involved. WIN5-RB has two types of layers with two different colors. (a) Conv+BN+ReLU \cite{nair2010rectified}: for layers 1 to $L-1$, BN is added between Conv and ReLU \cite{nair2010rectified}. (b) Conv+BN: for the last layer,
$K=1$ filters of size $F=7\times7$ is used to reconstruct the \(R(y)\approx\displaystyle -n\). In addition, a shortcut skip connecting the input (data layer) with the output (last layer) is added to merge the input data with \(R(y)\) as the final recovered image. 

\textbf{Regularization of BN.} Batch Normalization (BN) has a regularizing effect of improving the generalization of a learned model, which is motivated by the fact that data whitening improves performance. Particularly, this whitening process performs a linear transformation applied to the convolutions of the bottom layers before feeding into the top layers (ReLU \cite{nair2010rectified}). In WIN5 and WIN5-R without BN, the nonlinear transformation of $L^{th}$ layer is able to be formalized as $T_{n}(O_{L-1},\Theta )$, where $O_{L-1}$ is the output of the $(L-1)^{th}$ layer (Conv), nevertheless, with BN in WIN5-RB, which is changed to be 
\begin{equation} \label{eq:2}
T_{n}(BN(O_{L-1}),\Theta)
\end{equation} 
where BN is divided in two sub-operations: the first sub-operation normalizes the output of the bottom layer (Conv or ReLU \cite{nair2010rectified}), dimension-wise with zero mean and unit variance within a batch of training images; the second sub-operation optimally shifts and scales these normalized activations.
The learned parameters (means, variances, scaling and shifting) involved in the two sub-operations during training are utilized to infer \(R(y)\) during testing. 

\textbf{``Wider'' BN.} The integration of BN \cite{ioffe2015batch} into more filters will
further preserve the prior information of the training set. Actually, a number of state-of-the-art studies \cite{elad2006image, joshi2009image,xu2015patch} have adopted image priors (e.g. distribution statistic information) to achieve impressive performance. In our proposed models, the precomputed normalization parameters (means and variances) are used along with other trained network parameters to predict the distribution of \(R(y)\). The sparse distribution statistics performed by convolution and ReLU \cite{nair2010rectified} layers are updated during the training process. Eventually, their effects are embedded into these learned
normalization parameters that are simply a linear transformation applied to each activation. They transformation can be merged with respectively trained scaling or shifting parameters after the training of the network. Formally,
we can present the two sub-operations during test as
\begin{equation} \label{eq:3}
 \hat{O}_{L,K}=\frac{O_{L-1,K}-\mu_{L,K}}{\sqrt{s^2_{L,K}+\varepsilon}}
\end{equation}
\begin{equation} \label{eq:4}
  {BN}_{L,K}(\hat{O}_{L,K})=\gamma_{L,K} \hat{O}_{L,K}+\beta_{L,K} 
\end{equation}
where $\hat{O}_{L,K}$ is the normalized output of the convolution of the $K^{th}$ filter in $(L-1)^{th}$ layer (Conv) using the corresponding mean $\mu_{L,L}$ and variance $s^2_{L,K}$ of training set in the $L^{th}$ layer (BN); $\gamma_{L,K}$ and $\beta_{L,K}$ preserve the scaling and shifting parameters for correction after the normalization by learning from the training stage. They perform a linear transformation applied to the normalized output of each convolution layer. As a result, we can see in Fig.~\ref{fig:parameters}(d) that the generalization capability improved by ``wider'' BN can be observed distinctly by comparison experiments between WIN5-R (Blue line) and WIN5-RB (Red line).   

In this work, we employ Batch Normalization (BN) and residual learning (skip-connection) 
mostly for extracting pixel-distribution
statistic features and reserving training data means and variances in networks for denoising 
inference instead of  using the regularizing effect of improving the generalization of a learned model. In Fig.\ref{fig:workflow}, we illuminate the process of denoising inference by sparse distribution statistics features. We can consider BN as a cache area in WINs. Learned priors are preserved in WINs as knowledge base for denoising inference. When WIN has more channels to preserve more data means and variances, various combinations of these feature maps can corporate with residual learning to infer the noise-free images more accurately. 

\subsection{Implementation Details}
We implement the training with step learning rate policy along with basic learning rate 0.1. The stochastic gradient descent algorithm (SGD) with momentum 0.9 is adopted. Meantime, weight decay $1e{-4}$ and clip gradient 0.1 also are utilized to optimise training process. The batch size is deployed as 64 to balance the BN performance and training time. 


\section{Experiments}
\subsection{Datasets for Training and Testing}
\textbf{Training dataset.} 
We follow the experimental setup of RED-Net \cite{mao2016image} and 
use the BSD200-train (200 images) of the BSD500 dataset \cite{MartinFTM01} 
as our base training set and the BSD100-val of the ~\cite{MartinFTM01} is used for validation.
In addition, data augmentation (rotation or flip) is used to increase the sample size \textit{only}
for blind denoising model-WIN5-RB-B. 
We follow \cite{Kim_2016_VDSR,zhang2016beyond} to set the patch size as $41\times41$, and
crop 1,239,808 patches with the stride of 14 to train the model. We use the \textit{randn} function of the
internal MATLAB functions to add AWGN with different noise levels, i.e., $\sigma =30,50,70$.

\begin{figure}[ht]
\vspace{-1em}
\scriptsize
\centering
  \includegraphics[width=\textwidth]{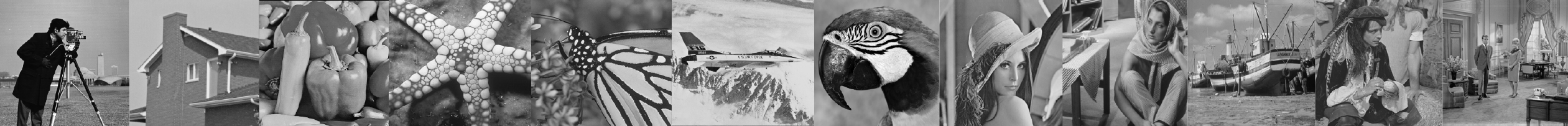}
  \caption{ The 12 widely used testing images (Set12). 
  }
   \vspace{-1em}
\label{fig:set12}
\end{figure}

\textbf{Test dataset.} We use BSD200-test~\cite{MartinFTM01} and the 12 standard test images, termed Set12, as shown in Fig.~\ref{fig:set12} for the evaluation. As there are various versions of Set12, we choose to resize the 12 images to be $481\times321$, same as the size of the majority of the images in training set (BSD200-train) \cite{MartinFTM01}.

 \begin{table}[!]
\scriptsize
\caption{\small The average results of PSNR (dB) / SSIM / Run Time (seconds) of different methods on the BSD200-test~\cite{MartinFTM01} (200 images). Note: WIN5-RB-B (blind denoising)  is trained on larger number of patches as data augmentation is adapted.This is the reason why WIN5-RB-B (trained on $\sigma=[0-70]$) can outperform WIN5-RB-S (trained on single $\sigma=10,30, 50, 70$ separately) in some cases.}
\label{tab:BSD-test}
\centering
  \begin{tabular}{llllllll}
    \toprule
   \multicolumn{8}{c}{PSNR (dB) / SSIM }                   \\
    \cmidrule{1-8}
    $\sigma$ & BM3D \cite{dabov2009bm3d}   & RED-Net \cite{mao2016image} & DnCNN \cite{zhang2016beyond} & WIN5 & WIN5-R & WIN5-RB-S & WIN5-RB-B\\
     
    \midrule
    10  & 34.02/0.9182 & 32.96/0.8963 & 34.60/0.9283  & 34.10/0.9205  & 34.43/0.9243  &  \textbf{35.83/0.9494}  & 35.43/0.9461 \\
    30  & 28.57/0.7823 & 29.05/0.8049 & 29.13/0.8060          &28.93/0.7987 & 30.94/0.8644 & \textbf{33.62/0.9193}  & 33.27/0.9263    \\
    50  & 26.44/0.7028 & 26.88/0.7230 & 26.99/0.7289          &28.57/0.7979 & 29.38/0.8251 &31.79/0.8831 & \textbf{32.18/0.9136}\\
    70  & 25.23/0.6522 & 26.66/0.7108 & 25.65/0.6709          &27.98/0.7875 & 28.16/0.7784 & 30.34/0.8362  & \textbf{31.07/0.8962}\\
    \bottomrule
      
   \multicolumn{8}{c}{Run Time(s)}                   \\
 
   \midrule
   
    30  &\textbf{1.67} &69.25  & 13.61 & 15.36 & 15.78 &  20.39  &  15.82 \\
    50  &\textbf{2.87} &70.34  & 13.76 & 16.70 & 22.72 &  21.79  &  13.79    \\
    70  &\textbf{2.93} &69.99  & 12.88 & 16.10 & 19.28 &  20.86  &  13.17   \\
    \bottomrule
  \end{tabular}
   \vspace{-2em}
\end{table}

     
  

\begin{table}[!]
\scriptsize
  \caption{\small The average PSNR(dB) / SSIM / Run Time (seconds)  of different
  methods on the 12 standard test images. WIN5-RB-B (blind denoising) is trained 
  for $\sigma=[0-70]$. The best results are highlighted in bold.}
  \label{tab:Set12-test}
  \centering
  \begin{tabular}{llllllll}
    \toprule
   \multicolumn{8}{c}{PSNR (dB) / SSIM }                   \\
    \cmidrule{1-8}
    $\sigma$ & BM3D \cite{dabov2009bm3d}   & RED-Net \cite{mao2016image} & DnCNN \cite{zhang2016beyond} & WIN5 & WIN5-R & WIN5-RB-S & WIN5-RB-B\\
     
    \midrule
    30  & 30.41/0.8553  & 30.48/0.8610       & 30.65/0.8644  & 30.42/0.8592  &  33.35/0.9142  &  \textbf{36.96/0.9495} & 35.83/0.9460   \\
    50  & 27.92/0.7947  & 28.03/0.7988       & 28.18/0.8054  & 29.52/0.8376  &  31.65/0.8896  &  34.12/0.9074 & \textbf{34.13/0.9323}   \\
    70  & 26.32/0.7451  & 27.95/0.7950       & 26.52/0.7546  & 28.89/0.8276  &  30.09/0.8529  &  32.32/0.8689 & \textbf{32.52/0.9145}    \\
    \bottomrule
  
  \multicolumn{8}{c}{Run Time(s)}                   \\
 
   \midrule
   
    30  &\textbf{1.47}  &62.71  & 9.24 & 14.39 & 13.58 &  14.76  &  16.95 \\
    50  &\textbf{2.40} &63.66  & 9.83 & 14.82 & 14.52 &  15.38  &  15.71   \\ 
    70  &\textbf{2.30} &63.20  & 9.42 & 14.55 & 13.86 &  15.41  &  13.56   \\
    \bottomrule

  \end{tabular}
\end{table}

\begin{figure*}[!]
 \vspace{-1em}
\centering
  \includegraphics[width=\textwidth]{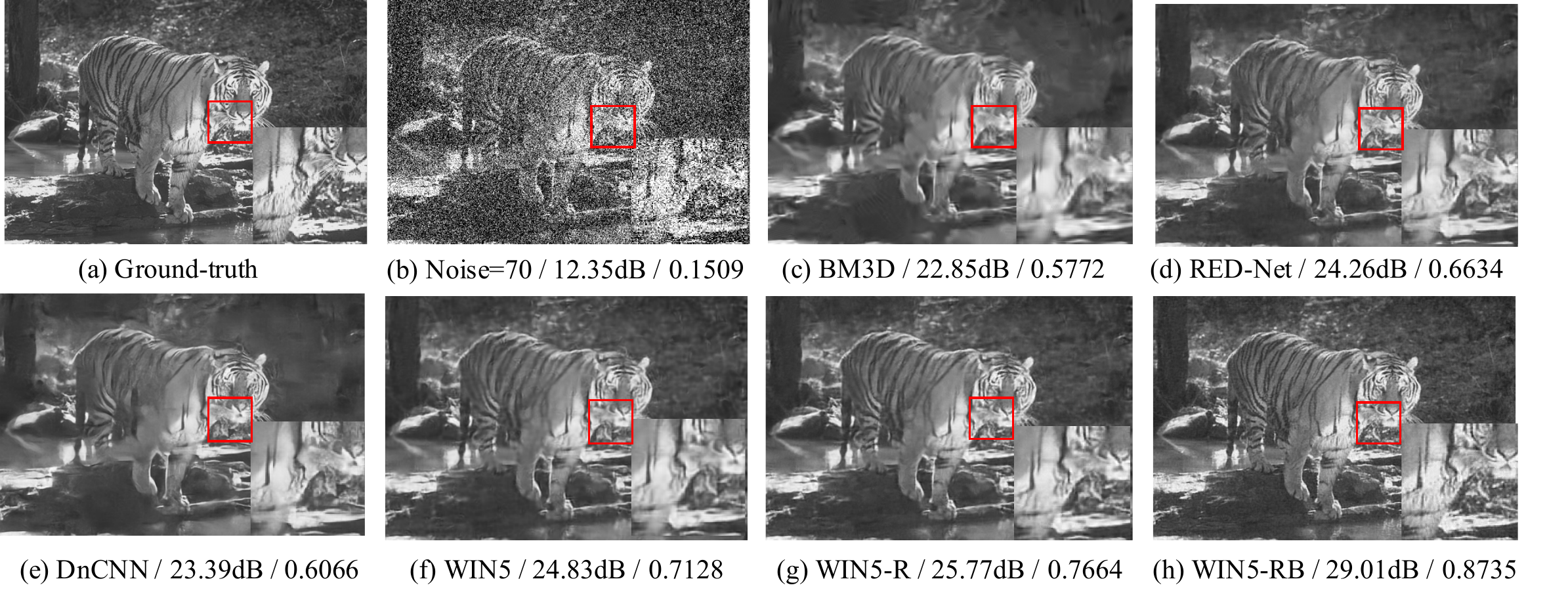}
  \caption{\small Visual results of one image from BSD200-test with $\sigma=70$
  along with PSNR(dB) / SSIM.}
   \vspace{-0.8em}
\label{fig:visual}
\end{figure*}

\subsection{Comparison with the State-of-the-Art}
We compare the proposed WIN5, WIN5-R and WIN5-RB methods with BM3D \cite{dabov2009bm3d},
DnCNN \cite{zhang2016beyond}, RED-Net \cite{mao2016image} for Gaussian denoising with $\sigma =30, 50, 70$.
BM3D \cite{dabov2009bm3d} is non-CNN based while both DnCNN \cite{zhang2016beyond}
and RED-Net \cite{mao2016image} are deep CNN-based
methods. The implementation code is either downloaded from the authors' websites or
implemented by our own with comparable or better performance than the official results.

\textbf{Quantitative Evaluation.} We evaluate our models through single
noise level-S (known noise level) and blind denoising-B (unknown noise level).
The average of PSNR/SSIM results of different methods on the BSD200-test
dataset and Set12 are shown in Table~\ref{tab:BSD-test} and Table~\ref{tab:Set12-test}.
As one can see, nearly all of our proposed models achieve the best results.
Compared to the best performance of existing methods (RED-NET \cite{mao2016image}
or DnCNN \cite{zhang2016beyond}), on BSD200-test and Set12, the plain network WIN5 outperforms
both methods at noise levels of 50 and 70 and has comparable performance at noise
level of 30; WIN5-R is able to obtain remarkable PSNR gain of 1.81 / 2.39 / 1.5 dB on BSD200-test and 2.7 / 3.47 / 2.14 dB on Set12, at noise levels of 30 / 50 / 70 respectively; WIN5-RB-S can yield exceptional results
with respectively 4.49 / 4.8 / 3.68 dB on BSD200-test and 6.31 / 5.94 / 5.8 dB gain on Set12 at the three noise levels. Note that this is the fist time that CNN-based denoising model can outperform other existing methods by more than 2 dB, or even 6 dB. 
It benefits from wider inference architecture 
capturing pixel distribution 
through wide reception fields.
The computational cost of our proposed WIN models are comparable to DnCNN while four times faster than RED-Net.

\textbf{Qualitative Evaluation.}
Fig.~\ref{fig:visual} illustrates the visual results of different methods. BM3D \cite{dabov2009bm3d} (non-CNN based method) tends to produce over-smoothed edges and textures; DnCNN \cite{zhang2016beyond} and RED-NET's \cite{mao2016image} outputs are better than BM3D \cite{dabov2009bm3d}; our proposed model, especially, WIN5-RB, with residual learning which helps to preserve original details by adding them back to the output by a skip connection, can yield more natural and accurate details in the texture as well as visually pleasant results.
\subsection{Blind denoising and Robustness}
\begin{wrapfigure}{r}{0.3\textwidth}
\vspace{-1em}
 \footnotesize
\centering
  \includegraphics[width=0.28\textwidth]{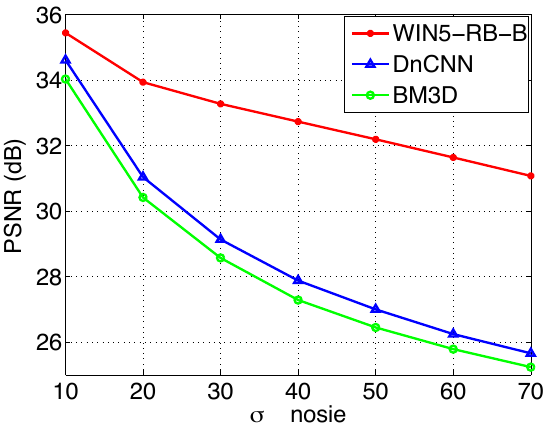}
  \caption{\small
  Behavior at different noise levels of average PSNR on BSD200-test. WIN5-RB-B (blind denoising) is 
  trained for $\sigma=[0-70]$. 
  }
\vspace{-1em}
\label{fig:behavior-blind}
\end{wrapfigure}
BSD200-test images now are corrupted with AWGN with different values of $\sigma=[0-70]$. WIN5-RB-B is trained on larger number
of patches as data augmentation is adapted. 
This is the reason why WIN5-RB-B (trained on $\sigma=[0-70]$) can outperform WIN5-RB-S (trained on single $\sigma=10, 30, 50, 70$ separately) in some cases, which is shown in the last column of Table~\ref{tab:BSD-test}. 
The average PSNR results of denoising behavior on BSD200-test dataset are shown in Fig.~\ref{fig:behavior-blind}. WIN5-RB-B performs more stable and generalizes better even on  higher noise levels than BM3D~\cite{dabov2009bm3d} and DnCNN~\cite{zhang2016beyond}. 

Particularly, as the noise level is increasing, the performance gain of WIN5-RB-B is getting larger, while the performance gain of DnCNN comparing to BM3D is not changing much as the noise level is changing. Compared with WINs, DnCNN is composed of even more layers embedded with BN. This observation indicates that the performance gain achieved by WIN5-RB does not mostly come from BN's regularization effect but the pixel-distribution features learned and relevant priors such as means and variances reserved in WINs. Both Larger kernels and more channels can promote CNNs more likely to learn pixel-distribution features.
\section{Conclusion and Discussion}

In this work, we primarily argue that wider CNNs that consist of
increased size of receptive fields and a number of neurons in convolution layers 
are able to learn pixel distribution features more effectively, which leads to 
remarkably Gaussian denoising results and even exceed the state-of-the-art methods available with large performance gains. Hence, an innovative pathway of designing image denoising models may start from a plain wide yet shallow architecture integrated with regularization and learning strategy techniques such as batch normalization and residual learning. 
More importantly, an innovative investigation of regularization is to utilize wider batch normalization for boosting neural networks' memory capacity to preserve more pixel distribution priors (mean and variance), with the empirical evidence demonstrate the generalization of the Gaussian denoising models is further improved. There are a number of observations which are suggested by our results, as discussed below.

\begin{wrapfigure}{r}{0.3\textwidth}
\centering
 \vspace{-0em}
  \footnotesize
\includegraphics[width=0.28\textwidth]{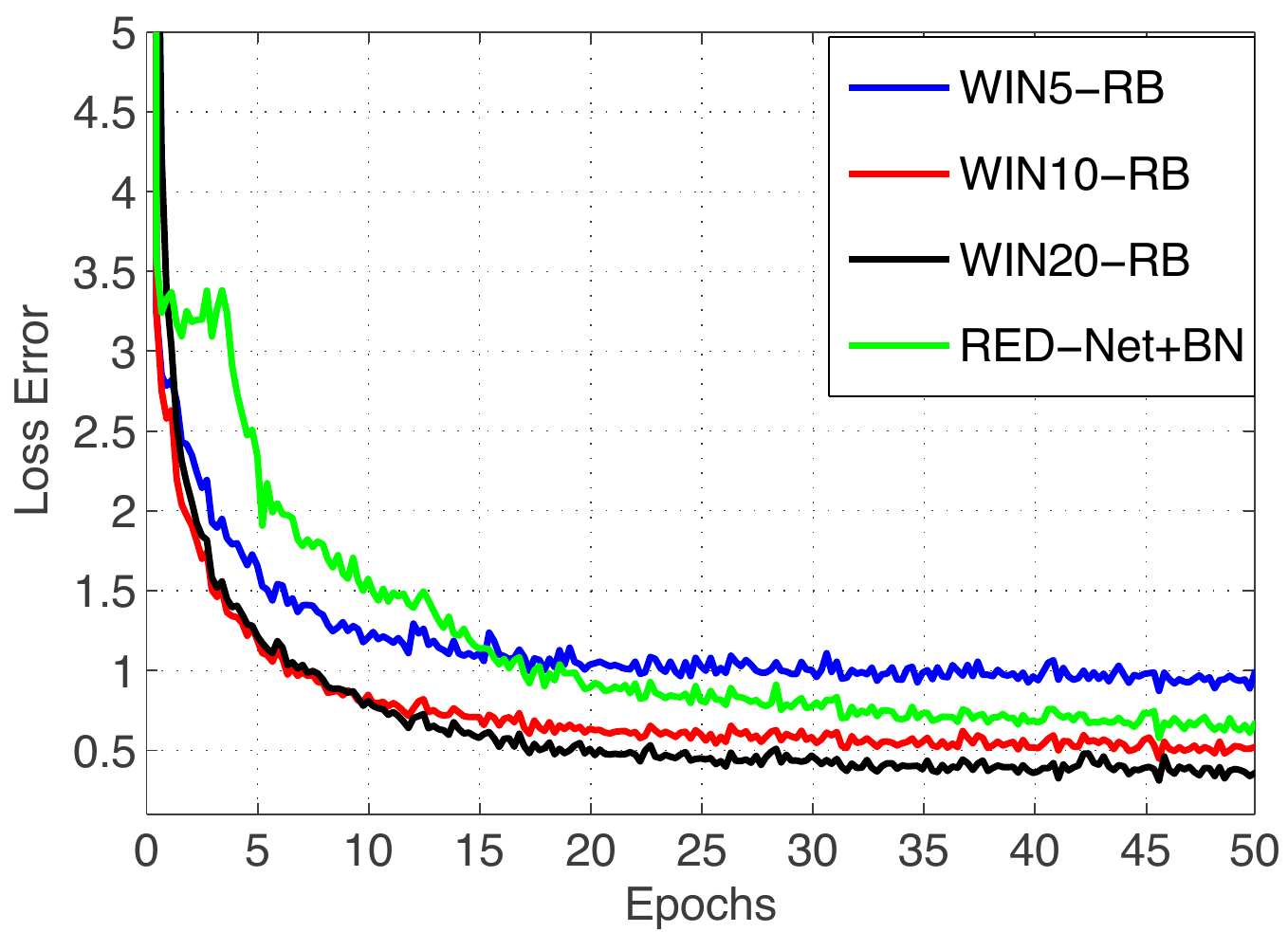}
\caption{\small Comparisons of loss errors of WIN-RBs at various depths
   as well as against RED-Net \cite{mao2016image} with BN added \cite{ioffe2015batch} layers on the same validation
   set. First, deeper is better as depth provides
   more non-linear representational compatibly. Second, ``Wider''
   BN \cite{ioffe2015batch} is able to enhance the network ``memory'' 
   to preserve more prior estimation. \textit{Note: Lower is better}. }
   \vspace{-0em}
\label{fig:deeper}
\end{wrapfigure}

\begin{wrapfigure}{r}{0.3\textwidth}
\centering
 \footnotesize
 \vspace{0em}
\includegraphics[width=0.28\textwidth]{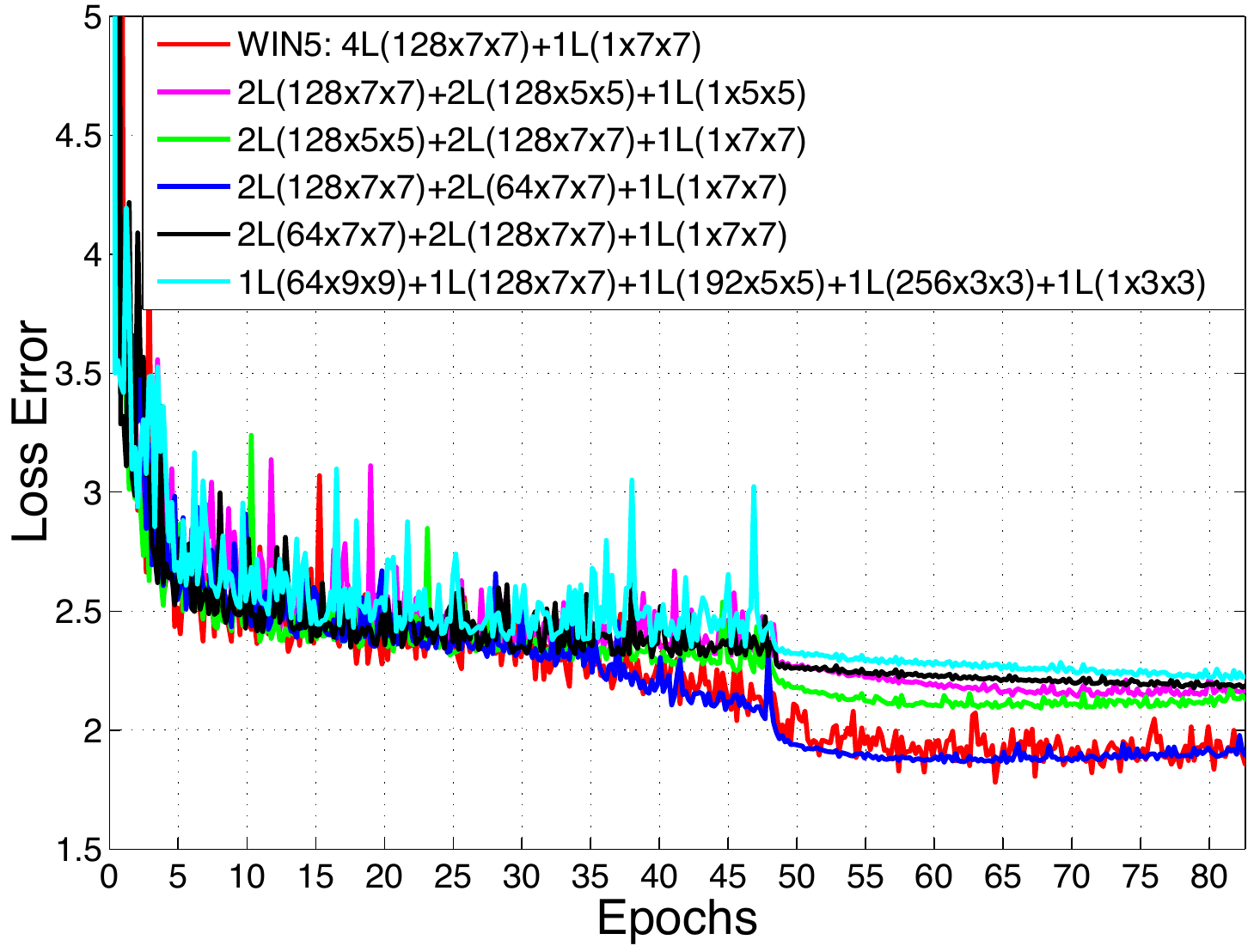}
\caption{\small Comparison of loss error of WIN5's variants. \textit{Note: Lower is better}. }
  \vspace{0em}
\label{fig:variousWINs}
\end{wrapfigure}

\textbf{Is deep CNN model necessary?} It depends on the task's complexity,
the training data available and the application scenarios (e.g., response time requirements).
Although the proposed wide inference network can achieve remarkable
results on Gaussian denoising, we still should notice the generalization
capability of CNN-based models can be improved by going deeper.
In our experiments, we compare the performance of 
the WIN-RBs with different depths (Fig.~\ref{fig:deeper}). The results indicate the performance also
can be further improved by increasing depths as long as degradation is reduced. 
\textbf{Investigation of WIN5 Variants:}
Fig. \ref{fig:variousWINs} shows the 
comparison of Gaussian denoising performance among WIN5's variants with different structures.
As one can see, the network (blue line) with the label  $2L(128\times7\times7)+2L(64\times7\times7)+1L(1\times7\times7)$ 
, which means each of the first 2 convolution layers
consists of 128 filters with size of $7\times7$, and both $3^{rd}$ and $4^{th}$
layers consist of 64 filters with size of $7\times7$, and the last convolution layer has 1 filters with size of $7\times7$, has the closest performance to WIN5 with $4L(128\times7\times7)+1L(1\times7\times7)$. 
To have the competitive performance with WIN5 while reducing model complexity, we may keep the size of filters in all layers to be $7\times7$ and decrease the number of filters after $2^{nd}$ layer. The key to WIN5's success is to embed larger size of filters in all convolution layers with at least 128 filters in the first two convolution layers. In CNNs, larger receptive fields may capture more pixel distribution statistics that shall be fed into convolutions for learning sparse features. Especially, a number of low-level features that are essential for denoising are primarily convolved in the first two layers, which need more neurons focusing on receptive regions simultaneously 
to extract plentiful significant sparse features.

\textbf{Prior and learning structure:} 
Designing a learning structure that remembers more prior information during training
can significantly boost the accuracy of estimation and inference. Prior has played a critical role in some early works. The well-known
Gaussian scale mixture (GSM) \cite{andrews1974scale} model employs a known multi-scale
wavelet representation as a prior to represent images statistics. In addition,
using Markov random fields (MRFs) \cite{rozanov1982markov} to define a prior over the image space is another approach to capturing
statistical regularities of pixel intensities. In contrast, CNNs have a greater representational power to learn priors from the training set, which is not only associated with the regression statistical framework~\cite{jain2009natural} but also able to benefit from network optimization techniques. Compared with other CNN-based models, in some sense, our proposed model employs a regularization technique--BN \cite{ioffe2015batch}, a learning strategy-Residual net \cite{kiku2013residual}, and degradation optimization skill-skip connections as an associative memory~\cite{hopfield1982neural} to preserve more statistical priors. Hence, our proposed WIN model can achieve remarkable performance boost.

\textbf{Contributions and further novelty:} In this work, we explore a novel strategy with CNNs rather than only a new structure of CNNs to solve a specific problem: removing Gaussian noise from images.
We demonstrate our key contributions as below: (1). We reveal that with increasing kernel size and channel number, CNNs prefer to learn similar pixel-distribution features, which exactly is a property of additive white Gaussian noise (AWGN). We call this property as "prior". (2). We demonstrate a new learning strategy by taking more consideration of the properties behind data-self rather than CNN-self only. This point may guide us to explore wider low-level vision tasks. (3). A (Prior+CNN)-based approach requires less training samples. Experimental results show that even the proposed WINs is trained with fewer samples but still perform much better than the CNN-based state-of-the-art methods, such as DnCNN is trained on 400 images applied with data-augmentation, and WIN5/WIN5-R/WIN5-RB (except for WIN5-RB-B) are trained on 200 images without data-augmentation.  This point may give us a new way to train a high-effective learning models in fewer training samples. (4). This work may prompt us to further explore the different effects and contributions of CNN's width and depth inference.  Our experiments show that the inference performance based CNN’s width (shallow but wider) is largely related to training data’s pixel-distribution features, but the one based CNN’s depth (narrow but deeper) mostly comes from non-linear reasoning.  

{\small
\bibliographystyle{ieee}
\bibliography{refer_WIN}
}
\newpage

\section*{I. Prior: pixel-distribution features} 

\begin{figure}[ht]
\centering
  \includegraphics[width=\textwidth]{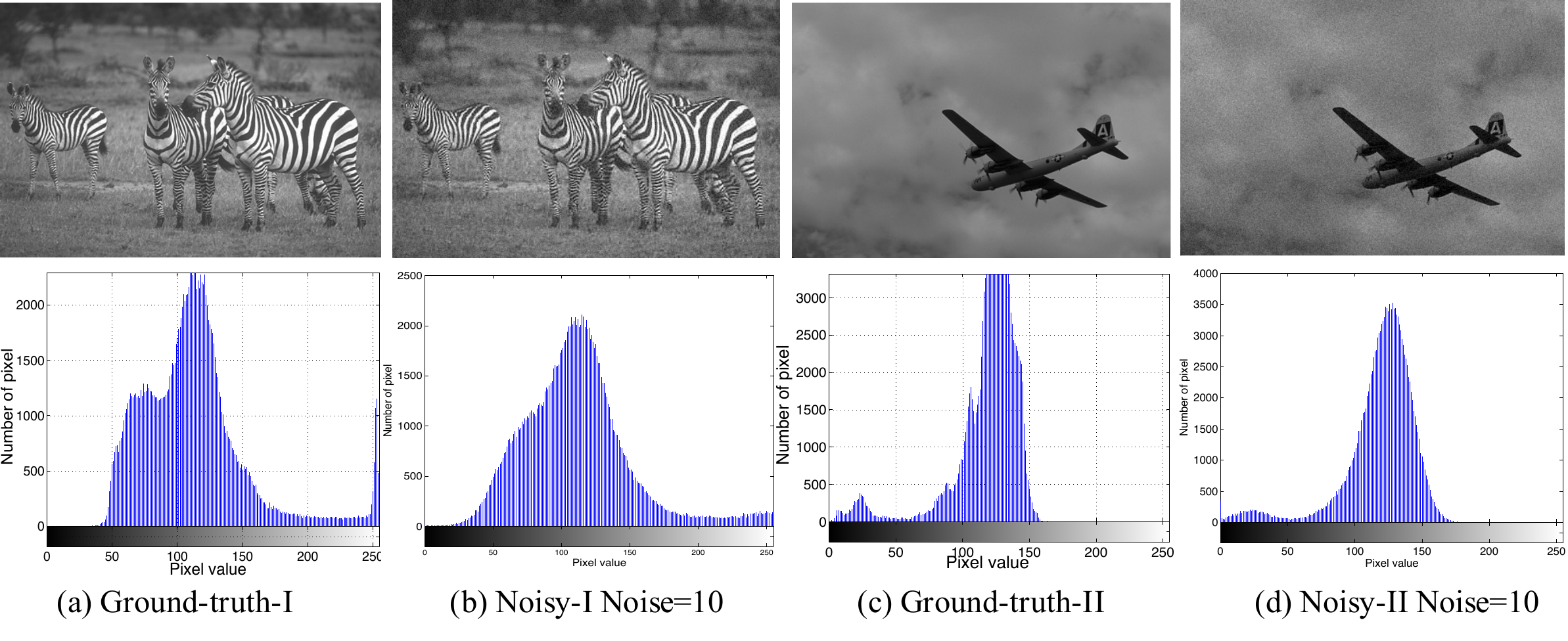}
  \caption{Compare the pixel distributions of histograms of two different
  images added additive white Gaussian noise (AWGN) with same noise level $\sigma=10$. }
\label{fig:histogram_N10}
\end{figure}

By comparing Fig.~\ref{fig:histogram_N10} and Fig.~\ref{fig:histogram_N50}, as one can see, 
the pixel-distribution in noisy images is more similar in higher noise level $\sigma=50$ than lower noise level $\sigma=10$. WIN infers noise-free images based on the learned pixel-distribution features, and it is easily to 
see that the higher the noise level is the more similar the pixel-distribution features are. Thus, WIN can learn more pixel-distribution features from noisy images having higher level noise. This is the reason why WIN performs even better in higher-level noise (see Table\ref{tab:BSD-test}). WIN learns the similar pixel-distribution features, and we call it as “Prior”, which gives much contribution for performance. 

Moreover, In Table\ref{tab:BSD-test}, the WIN5-RB-B is trained on more samples that are generated with data-augmentation skill. As one can see, the running time is much better than WIN5-RB without data-augmentation. This result indicates more training samples can help WINs learn more accurate similar features that can accelerate the inference.

\section*{II.Having Knowledge Base with Batch-Normal}
In this work, \textit{we employ Batch Normalization (BN) and residual learning (skip-connection) 
mostly for extracting pixel-distribution
statistic features and reserving training data means and variances in networks for denoising 
inference instead of 
using the regularizing effect of improving the generalization of a learned model.}

First, the input-to-output skip-connection guides WINs to infer the opposite noise,
which always follows consistent distribution. Second, 
the regularizer-BN can keep the data distribution the same as input: Gaussian distribution. 
This distribution consistency between input and regularizer ensures more pixel-distribution
statistic features can be extracted accurately. 
The integration of BN \cite{ioffe2015batch} into more filters will
further preserve the prior information of the training set. Actually, a number of state-of-the-art studies \cite{elad2006image, joshi2009image,xu2015patch}
have adopted image priors (e.g. distribution statistic information) to achieve impressive performance. 
\section*{III. Can Batch Normalization work without a skip connection?} 
\begin{wrapfigure}{r}{0.4\textwidth}
\vspace{-0em}
  \centering 
    \includegraphics[width=0.38\textwidth]{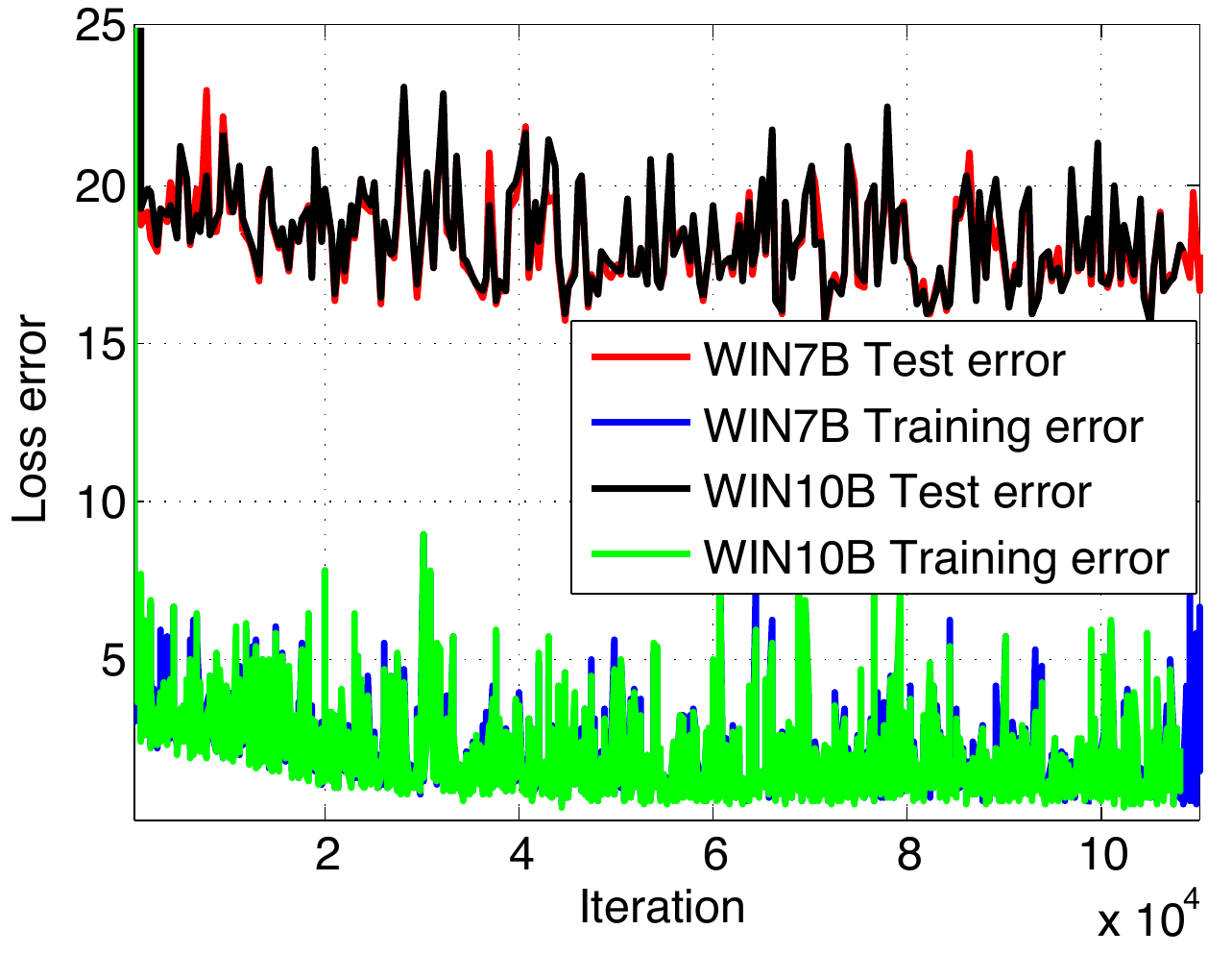}
  \caption{Comparing the both training and validation-test loss error during training between WIN7+BN and WIN10+BN.}
  \label{fig:WINB_losserror}
 \vspace{-1em}
\end{wrapfigure}

WIN+BN cannot work without the input-to-output skip connection and is always over-fitting. 
In Fig.\ref{fig:WINB_losserror}, as one can see, both deeper WINs+BN composed of 7 and 10 Conv+ReLU+BN layers are over-fitting without skip-connection's assistance.
In WIN5-RB's training, BN keeps the distribution of input data consistent and the skip connection
can not only introduce residual learning but also guide the network to extract the certain features in common: pixel-distribution. 
Without the input data as a comparison, BN could bring negative effects as keeping the each input distribution same, especially, when a task is to output pixel-level feature map. 
In DnCNN, two BN layers are removed from the first and last layers,
by which a certain degree of the BN's negative effects can be reduced.
Meantime DnCNN also highlights network's generalization ability largely
relies on the depth of networks.

\section*{IV. More Visual Results} \label{morevisual}
More visual results are essential evidence to show the effectiveness and the advantages of our methods. 
We have various images from two different datasets,
BSD200-test and Set12, with noise levels $\sigma=10, 30, 50, 70$ applied separately.

\textbf{One image from BSD200-test with noise level=10 }

\begin{figure}[ht]
\centering
  \includegraphics[width=\textwidth]{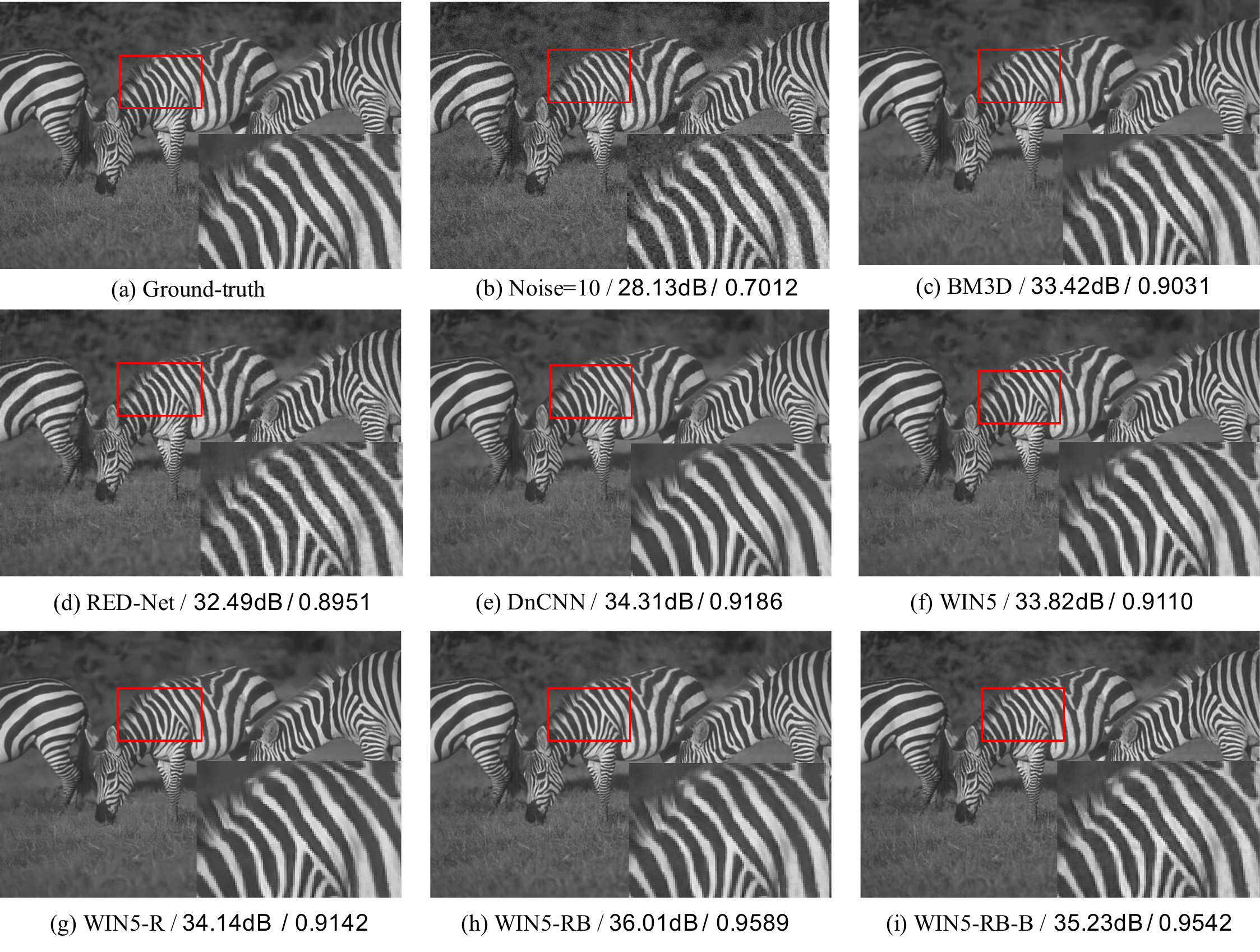}
  \caption{Visual results of one image from BSD200-test with noise level $\sigma=10$ 
  along with PSNR(dB) / SSIM. As we can see, our proposed methods 
  can yield more natural and accurate details 
  in the texture as well as visually pleasant results.}
\label{fig:visual00}
\end{figure}
\newpage



\textbf{Comparing 7x7 filter-size WINs with 13x13 filter-size WINs for noise level=30}
\begin{figure}[ht]
\centering
  \includegraphics[width=\textwidth]{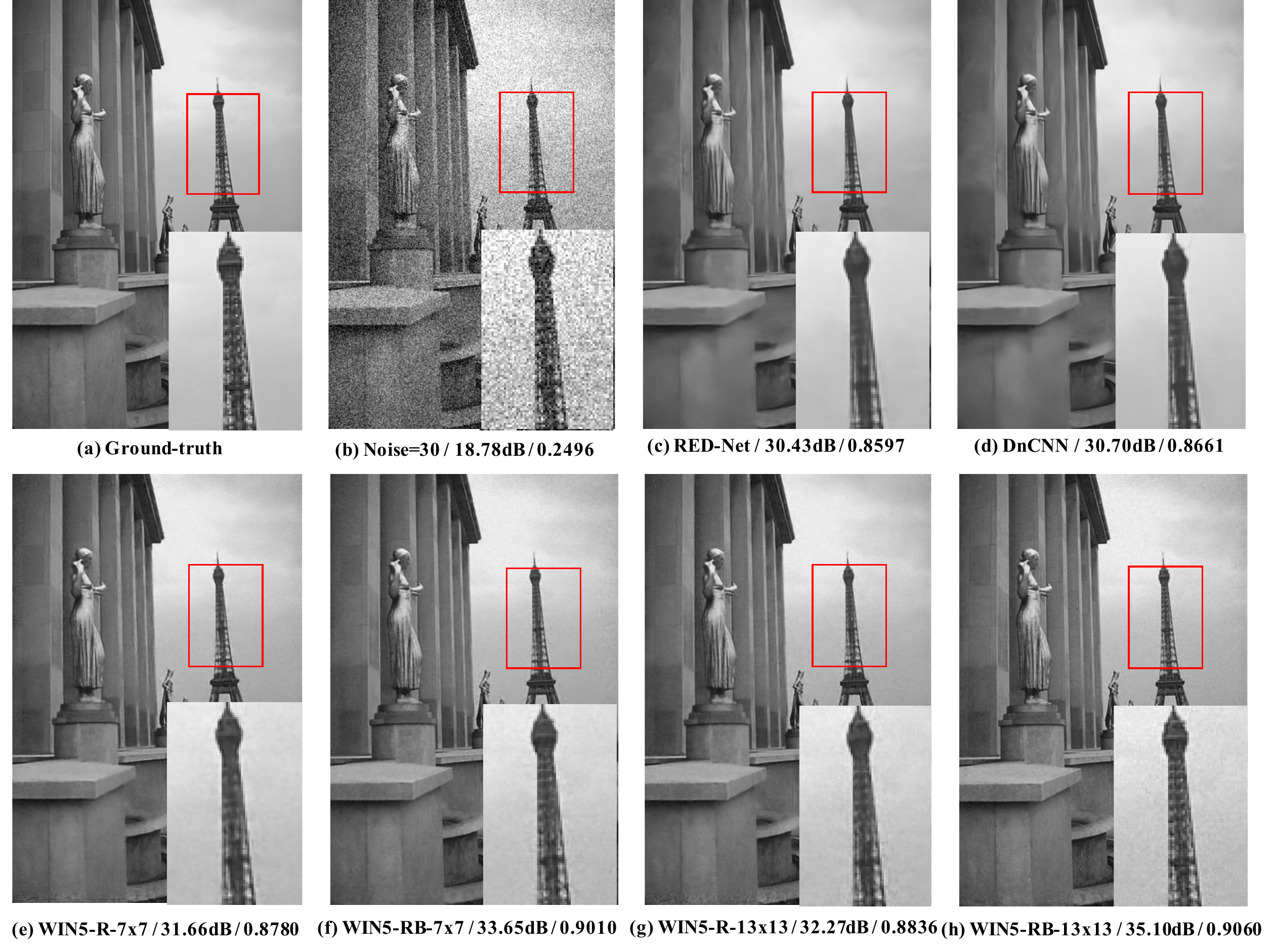}
  \caption{Comparing various WINs with different size of filters: Visual results of one image from BSD200-test with noise level $\sigma=30$ 
  along with PSNR(dB) / SSIM. As we can see, Increasing filter size can further improve performance.}
\label{fig:visual5}
\end{figure}
\newpage

\textbf{One image from BSD200-test with noise level=50}
\begin{figure}[ht]
\centering
  \includegraphics[width=\textwidth]{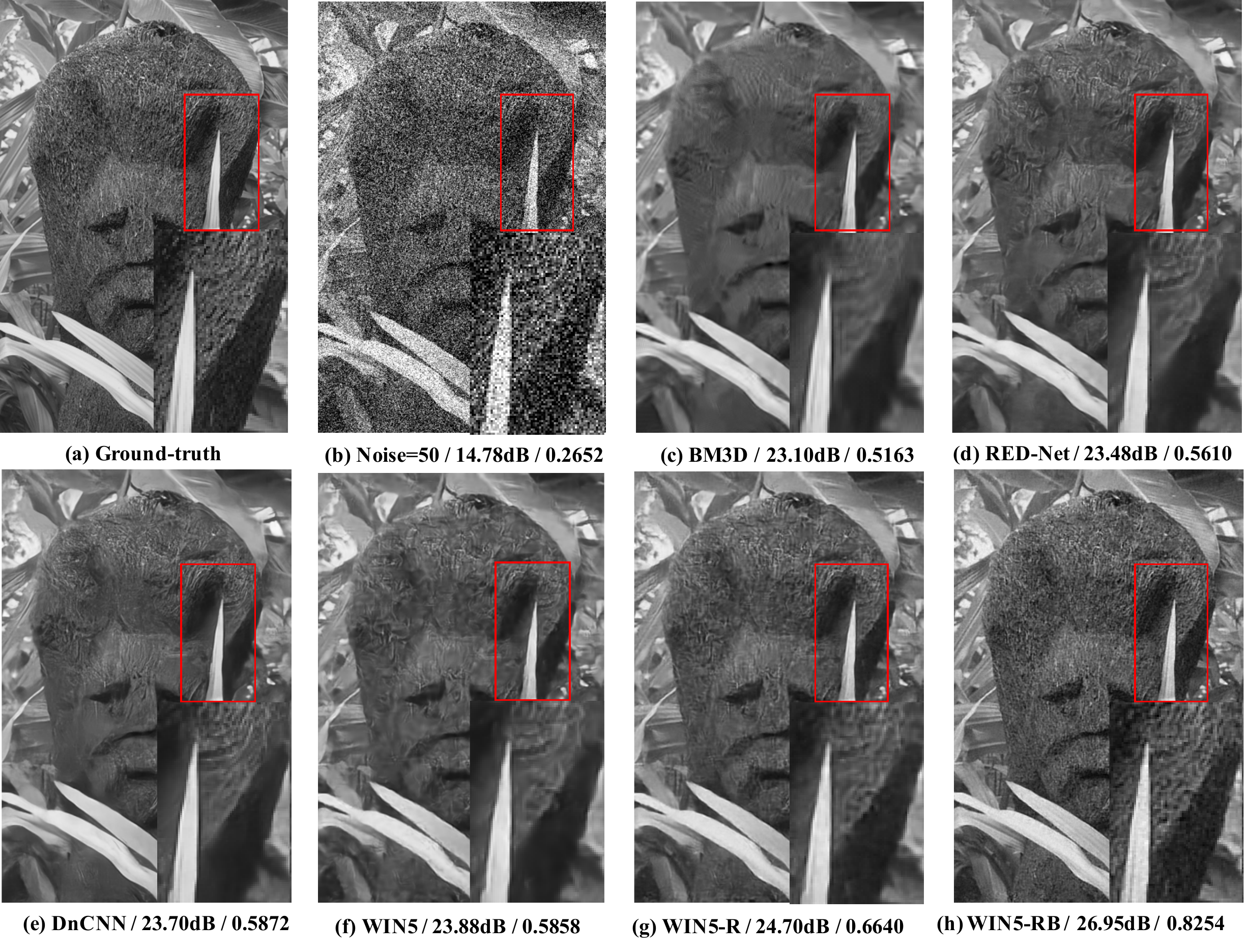}
  \caption{Visual results of one image from BSD200-test with noise level $\sigma=50$ 
  along with PSNR(dB) / SSIM. As we can see, our proposed methods 
  can yield more natural and accurate details 
  in the texture as well as visually pleasant results.}
\label{fig:visual2}
\end{figure}

\newpage

\textbf{One image from BSD200-test with noise level=70 }
\begin{figure}[ht]
\centering
  \includegraphics[width=\textwidth]{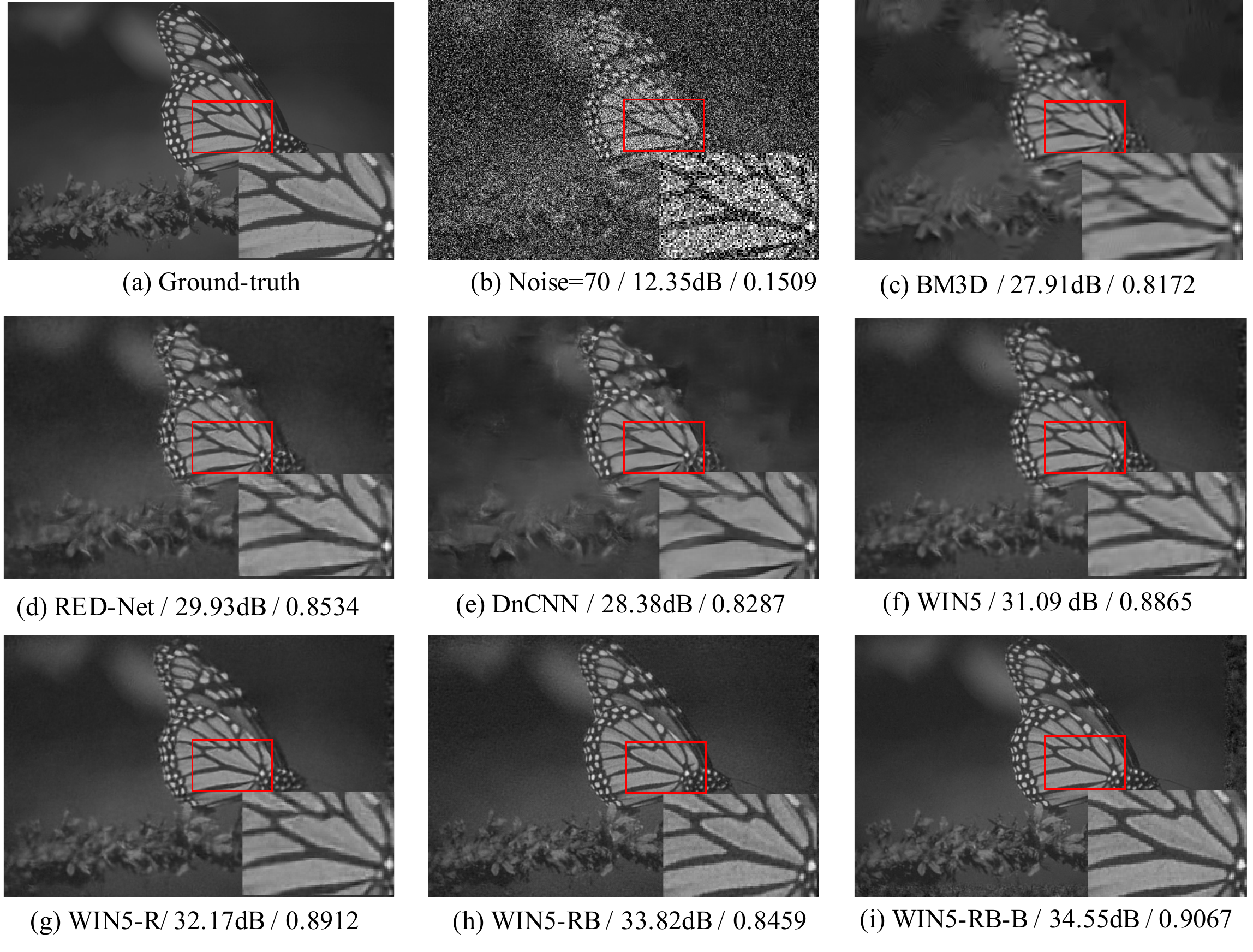}
  \caption{Visual results of one image from BSD200-test with noise level $\sigma=70$ 
  along with PSNR(dB) / SSIM. As we can see, our proposed methods 
  can yield more natural and accurate details 
  in the texture as well as visually pleasant results.}
\label{fig:visual1}
\end{figure}

\newpage




\textbf{Comparing 7x7 filter-size WINs with 13x13 filter-size WINs for noise level=30}

\begin{figure}[ht]
\centering
  \includegraphics[width=\textwidth]{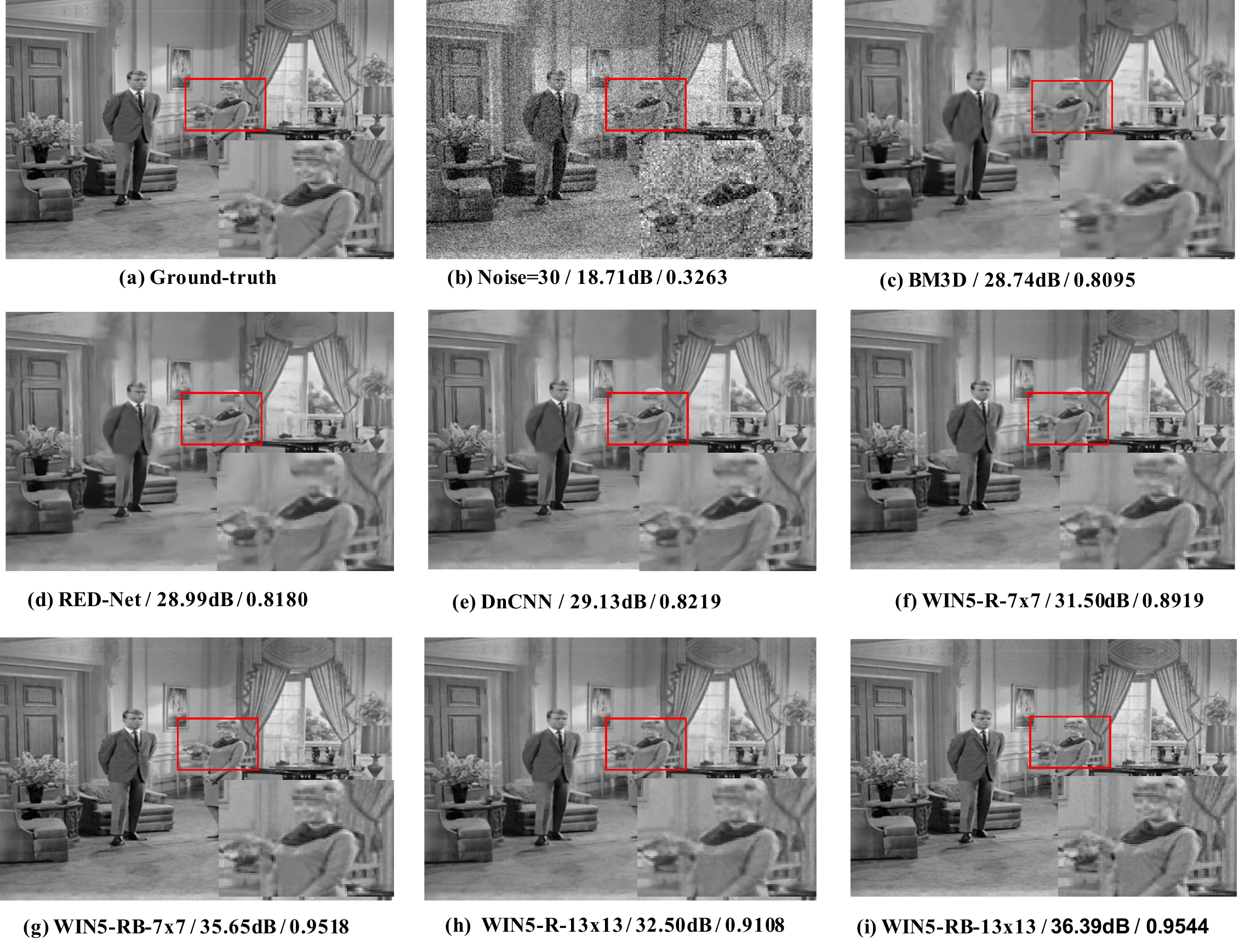}
  \caption{Visual results of one image from Set12 with noise level $\sigma=30$ 
  along with PSNR(dB) / SSIM. As we can see, our proposed methods 
  can yield more natural and accurate details 
  in the texture as well as visually pleasant results.}
\label{fig:visual6}
\end{figure}

\end{document}